# Critical Appraisal of Fairness Metrics in Clinical Predictive AI

João Matos [1], Ben Van Calster [2,3], Leo Anthony Celi [4,5,6], Paula Dhiman [1], Judy Wawira Gichoya [7], Richard D. Riley [8,9], Chris Russell [10], Sara Khalid [1], Gary S. Collins [1]

[1] Centre for Statistics in Medicine, Nuffield Department of Orthopaedics, Rheumatology and Musculoskeletal Sciences, University of Oxford, Oxford, UK
[2] Department of Development and Regeneration, KU Leuven, Leuven, Belgium
[3] Leuven Unit for Health Technology Assessment Research (LUHTAR), Leuven, Belgium
[4] Beth Israel Deaconess Medical Center, Boston, MA, USA
[5] Laboratory for Computational Physiology, Massachusetts Institute of Technology, Cambridge, MA, USA
[6] Department of Biostatistics, Harvard T H Chan School of Public Health, Boston, MA, USA
[7] Department of Radiology and Imaging Sciences, Emory University, Atlanta, GA, US
[8] Department of Applied Health Sciences, School of Health Sciences, College of Medicine and Health, University of Birmingham, Birmingham, UK
[9] National Institute for Health and Care Research (NIHR) Birmingham Biomedical Research Centre, Birmingham, UK
[10] Oxford Internet Institute, University of Oxford, Oxford, UK

## Corresponding Author

João Matos
Centre for Statistics in Medicine,
Nuffield Department of Orthopaedics, Rheumatology & Musculoskeletal Sciences,
University of Oxford,
Oxford,
OX3 7LD, United Kingdom.
Email: joao.matos@ndorms.ox.ac.uk

## Funding

JM is funded by a Clarendon Fund Scholarship at University of Oxford. GSC and RDR are supported by the EPSRC (Engineering and Physical Sciences Research Council) grant for "Artificial intelligence innovation to accelerate health research" (EP/Y018516/1), and MRC-NIHR Better Methods Better Research grant (MR/Z503873/1). RDR is supported by the National Institute for Health and Care Research (NIHR) Birmingham Biomedical Research Centre at the University Hospitals Birmingham NHS Foundation Trust and the University of Birmingham. GSC and RDR are NIHR Senior Investigators. The views expressed are those of the author(s) and not necessarily those of the NHS, the NIHR or the Department of Health and Social Care. LAC is funded by the National Institute of Health through R01 EB017205, DS-I Africa U54 TW012043-01 and Bridge2AI OT2OD032701, and the National Science Foundation through ITEST #2148451. JWG is a 2022 Robert Wood Johnson Foundation Harold Amos Medical Faculty Development Program and declares support from Lacuna Fund (#67), NHLBI Award Number R01HL167811 and NIH common fund award 1R25OD039834-01. BVC is supported by the Research Foundation - Flanders (FWO) grant G097322N, Kom Op Tegen Kanker grant 13583, Internal Funds KU Leuven grant C24M/20/064. SK receives funding from Wellcome Trust and UKRI. The funders had no role in considering the study design or in the collection, analysis, interpretation of data, writing of the report, or decision to submit the article for publication.

## Conflicts of Interest

No conflicts of interests with this specific work are declared.

## Contributions

JM and GSC conceived the study and this paper. JM conducted literature search, data extraction, and analysis. JM drafted the manuscript with input and edits from GSC. All authors were involved in revising the article critically for important intellectual content and approved the final version of the article. JM is the guarantor of this work. The corresponding author attests that all listed authors meet authorship criteria and that no others meeting the criteria have been omitted.

## Abstract

Predictive artificial intelligence (AI) offers an opportunity to improve clinical practice and patient outcomes, but risks perpetuating biases if fairness is inadequately addressed. However, the definition of "fairness" remains unclear. We conducted a scoping review to identify and critically appraise fairness metrics for clinical predictive AI. We defined a "fairness metric" as a measure quantifying whether a model discriminates (societally) against individuals or groups defined by sensitive attributes. We searched five databases (2014–2024), screening 820 records, to include 41 studies, and extracted 62 fairness metrics. Metrics were classified by performance-dependency, model output level, and base performance metric, revealing a fragmented landscape with limited clinical validation and overreliance on threshold-dependent measures. Eighteen metrics were explicitly developed for healthcare, including only one clinical utility metric. Our findings highlight conceptual challenges in defining and quantifying fairness and identify gaps in uncertainty quantification, intersectionality, and real-world applicability. Future work should prioritise clinically meaningful metrics.

## Introduction

Clinical prediction models are typically derived using regression or machine learning methods, collectively known as predictive artificial intelligence (AI)[1]. These models can be *diagnostic*, estimating the probability that an individual currently has a condition (typically a disease), or *prognostic*, estimating the likelihood of an individual developing a clinical outcome over a specific time period [2]. Predictive AI promises to improve patient outcomes and reduce costs to health systems by supporting clinical decision making and risk communication [3]. However, despite being abundant in the biomedical literature [4], their real-world impact remains limited [5], with a few exceptions (e.g., FRAX [6], QRISK3 [7]).

Several challenges hinder successful implementation of predictive AI, including long-standing issues with reporting quality [8,9] and transparency which compromise reproducibility and independent evaluation [10]. In response, the TRIPOD guideline was published in 2015 to provide minimum reporting recommendations [11]; these were updated to TRIPOD+AI in 2024 to encompass AI methods [12]. Further, design and methodological limitations (e.g., small sample sizes, increased risk of overfitting) affect robust model development [13–16], often resulting in poor or misleading model performance [17] and poor generalisation to new settings [18].

Algorithmic bias – which occurs *when data or analysis biases are encoded directly or indirectly into a model during its development* – adds a further layer of complexity to these challenges in evaluating a model's performance [19]. These biases can arise from unrepresentative data of target populations [20], different underlying disease distributions [21], existing health disparities [22], biases in medical devices [23,24], and other sources [13]. Such biases are inherently dependent on the notion of sensitive (or protected) attributes: *a characteristic, variable, dimension, or axis according to which fairness can be evaluated*. Protected characteristics can vary by region [25,26,27], and in medicine, some may reflect biological differences that legitimately influence health outcomes [28, 29, 28,30].

Obermeyer and colleagues' seminal work shed light on the issue of fairness in clinical predictive AI [22]. A model used to systematically predict patients' future health needs was found to underestimate the needs of Black patients in the US compared to White patients. The root cause of this bias was the algorithm's reliance on healthcare costs as a proxy for health needs, which failed to account for systemic disparities in access to care [22]. Deploying such models risks exacerbating existing health disparities in the care of individuals or groups of individuals [31], leading to "unfairness".

### Defining "fairness"

Evaluating the performance of a clinical prediction model (e.g.,statistical discrimination, calibration, or clinical utility; Box 1) typically focuses on the estimation at population level (i.e., averaged across all individuals), masking potential differential model behaviour within that population. However, model performance is expected to naturally vary across subgroups. It is therefore important to understand the nature and magnitude of any differential model behaviour during model evaluation, and signals to suggest "unfairness". This issue, known as hidden stratification, occurs *when a model appears to perform well at the population level but* *exhibits poorer performance in one or more subgroups*, potentially leading to disparities in its predictions [32,33].

When evaluating potentially "unfair"clinical prediction models, a fundamental question arises: how to define "fairness"? In the context of predictive AI, fairness has been conceptualised in various ways, including "group fairness" [34], fairness through "unawareness" [35] or "awareness" [36], "counterfactual fairness" [37], and "minimax fairness" [38], among others. Despite an abundance of fairness definitions (Supplementary Table 2.2), there remains a notable gap in a precise and widely accepted definition of a "fairness metric".

### Motivation and Aims

The limited understanding of fairness metrics hinders the development of specific definitions and recommendations in reporting guidelines, such as TRIPOD+AI[12], the FUTURE-AI consensus guideline for trustworthy and deployable AI in healthcare [39], and the STANDING Together consensus recommendations [31]. These guidelines address fairness definitions and corresponding metrics with caution and minimal specificity. While previous notable efforts have reviewed fairness in machine learning [40–45], they rarely focus on clinical prediction models nor provide comprehensive or critical perspective [46,47] (Supplementary Table 2.4).

We aimed to conduct a scoping review to identify and critically appraise key definitions and fairness metrics reported in the literature on clinical prediction models. Three questions were investigated: *(1)* Which fairness metrics have been proposed, applied, and analysed in the clinical predictive AI literature; *(2)* How should each metric be interpreted, in light of existing ethical and legal frameworks; *(3)* When is the use of each metric justified.

## Methods

We conducted a scoping review to identify and compile fairness metrics used in the clinical predictive AI literature, irrespective of the clinical domain or modelling approach used for the prediction task. To maximise coverage, our inclusion criteria and article selection process were intentionally broad. We followed the methodological framework by Arksey and O'Malley [48] and adhered to the Preferred Reporting Items for Systematic Reviews and Meta-Analyses extension for Scoping Reviews (PRISMA-ScR)[49].

We define a "fairness metric" as *a measure that quantifies the extent to which a model's output does or does not discriminate (in the societal sense, based on a given notion of fairness) against individuals or groups defined by a sensitive attribute* [50] (glossary in Box 1).

### Search Strategy

We searched literature published between 1 January 2014 and 22 October 2024. Searches were carried out (by JM) in five databases: PubMed, ACM Digital Library, IEEE Xplore, arXiv, and medRxiv. Four main concepts were searched for: "fairness", "metric", "clinical", and "model" (or equivalent terms), and we considered the full-text. The full search queries for each database are detailed in Supplementary Table 2.1. Grey literature was included through backward citation tracking and was conducted throughout the review, from 22 October 2024 until 1 May 2025.

## Eligibility criteria

Studies that directly engaged with fairness considerations in applied clinical prediction models were initially included without restrictions on how fairness was addressed. In parallel, backward citation searching was conducted to maximise the identification of relevant fairness metrics and to retrieve additional details on their definitions and use. This process allowed for the inclusion of both review articles and original works that proposed, defined, or applied fairness metrics, even if these metrics were not originally developed for healthcare contexts.

Studies were excluded if the reported "metrics" did not align with our definition of a fairness metric (Box 1). We excluded causal fairness notions – which require counterfactual predictions or additional steps. Finally, metrics lacking sufficient detail to clearly report their definition and operationalisation were excluded.

## Data Extraction

For each identified metric, the formula and any necessary information for its computation were extracted by JM. In case families of metrics were proposed (e.g., Equity-Scaled metric[51]), we reported specific examples as applicable (e.g., Equity-Scaled AUROC [51]). Metrics with ambiguous or unconventional names were standardised to ensure consistency in comparison with other metrics and alignment with the literature. For example, upon reviewing implementation details, the "Discrimination Index" was redefined as "F1-Score Parity"[52].

We categorised each metric based on its core attributes, including its proposed domain (healthcare or otherwise) and its classification within existing fairness taxonomies [1,36,44,53] (by JM). We also identified alternative names for metrics, their parent fairness metrics, and the required inputs for computation (e.g., type of model output, outcome labels, or predictor conditioning). Further, we examined the types of sensitive attributes each metric supports (binary, multi-group, or continuous). We assessed interpretation aspects, including "bias-preserving" versus "bias-transforming" properties [1,36,44,53], outcome prevalence dependency, and target fairness values.

## Taxonomy of fairness metrics

Multiple and conflicting fairness taxonomies (n = 12; Supplementary Table 2.3) exist in the literature. Figure 1 presents the adopted taxonomy for fairness metrics, which builds on previous taxonomies and is guided by three domains:

***1. Performance dependency:*** evaluation of fairness begins with considering if it should be evaluated with respect to model performance (performance-dependent fairness; "supervised"), or not (performance-independent fairness; "unsupervised"; with no consideration of "outcome labels" or "ground-truth labels"). This distinction hinges on the interpretation of fairness and validity of the labels used to develop and evaluate the model. If these outcome labels are accepted as a fair representation of the target population, a performance-dependent metric is justified (Box 3). Conversely, if the labels risk encoding health inequities that the model should not reinforce (e.g., disparities in disease prevalence, access to care, or historical disease patterns), performance-independent metrics could be more appropriate (Box 2). In practice, what is being enquired is whether we are satisfied with the *status quo* of the data and past outcome labels (Figure 1).

**2. *Level of model fairness:*** The next consideration is whether fairness is to be assessed at the level of estimated probability (using $\hat{p}$) or at the predicted class level ($\hat{Y}$, "classification"). Probability-based metrics (e.g., *mean score parity*) assess fairness without applying a decision threshold to the estimated probability. In contrast, threshold-dependent metrics (e.g., *statistical parity*), evaluate fairness at the class prediction level which depends on applying a threshold to convert the estimated probability into a class prediction. This distinction has also been described in the literature as "informing with a risk score" (probability-based), versus "decision support with a classification" (threshold-dependent) [54].

***3. Type of performance metric:*** If a performance-dependent fairness metric is chosen, the focus shifts to what type of model performance should be compared across groups.

**a. Probability-based** metrics focus on:

- **i. calibration**, ensuring estimated probabilities reflect actual event proportions (e.g., *calibration-in-the-large parity*);
- **ii. discrimination**, ensuring individuals who experience events receive higher probability estimates (e.g., *AUROC parity*);
- **iii. overall** performance (e.g., *Brier score parity*).

**b. Threshold-dependent** metrics focus on:

- **i. partial** metrics (e.g., *equality of opportunity difference,*);
- **ii. summary** metrics (e.g., *accuracy gap*);
- **iii. clinical utility** (e.g, *subgroup net benefit*).

Further, fairness metrics were categorised as *individual*[36] or *group* fairness[44] (Box 1).

## Qualitative Critical Appraisal

We developed a data extraction form (Supplementary Material 1) to critically appraise each metric assessing its suitability, limitations, and potential pitfalls in clinical scenarios. This included outlining its justified use and potential harm to patients when the metric is not satisfied. An overall critical appraisal and guidance is provided for different groups of fairness metrics. To aid interpretation, we categorised our recommendations into three levels of guidance: (1) *Recommended if* – metrics that are generally suitable and align with clinical and ethical standards under specific conditions; (2) *Use with caution* – metrics that may be relevant but require careful contextual consideration and are not essential; and (3) *Inadvisable* – metrics with substantial limitations or risk of harm that outweigh potential benefits in most scenarios.

# Results

A total of 927 records were identified through database searches. After removing duplicates and ineligible records based on language and format, 820 records proceeded to screening. 644 records were excluded for not focussing on clinical prediction models, not addressing fairness, or lacking fairness metrics. The remaining 175 reports were sought for retrieval, with 157 further excluded during data extraction due to the absence of fairness metric proposals, definitions, or applications. Ultimately, 19 reports were included in the review. An additional 22 reports were identified through backward citation searches, resulting in a total of 41 unique articles containing fairness metrics relevant to clinical prediction models (Supplementary Figure 2.1).

A total of 62 fairness metrics were identified from the 41 papers that met our definition of a fairness metric (Box 1). Metrics that did not meet this definition were excluded, even if they were labelled as such in the literature (e.g., *counterfactual fairness* [37], which is often referred to as a metric in subsequent papers, was not included in our analysis). A comprehensive and detailed formalisation (including mathematical formulation) of each metric can be found in the Supplemental Material 1.

Many fairness metrics originated from AI venues (n = 20/41) rather than biomedical or applied ethics research (n = 13/41 and n = 8/41, respectively). Only 18 metrics were explicitly defined for healthcare applications.

## Metrics found

### *Performance-independent metrics*

Among performance-independent metrics (n = 15)[41, 55, 56, 57, 58, 59, 60, 61, 62, 76], most were group fairness metrics (n = 13), with only two individual fairness metrics identified. These metrics were further divided into probability-based (n = 6, e.g., *mean score parity* [76]) and threshold-dependent (n = 7, e.g., *statistical parity* [60]). However, only 3 of these 16 metrics were proposed in healthcare applications.

### *Performance-dependent metrics*

Performance-dependent metrics (n = 47) [29, 34, 45, 51, 52, 56, 63, 64, 65, 66, 67, 68, 69, 70, 71, 72, 73, 74, 75, 76, 77, 78, 79, 80, 81, 82, 83, 84, 85, 86, 87, 88] were substantially more common, with all focusing on group fairness. These were divided into probability-based metrics (n = 21) and threshold-dependent metrics (n = 26). Among probability-based metrics, the most frequent were discrimination-based (e.g., *area under the receiver operator curve [AUROC] parity* [63], n = 11), ensuring the model assigns higher probabilities to individuals experiencing the event of interest, followed by calibration-based metrics (e.g., *expected calibration error parity* [68], n = 5) and overall metrics (e.g., *brier score parity* [64], n = 5). Threshold-dependent metrics primarily assessed fairness using confusion matrix-derived measures, with partial metrics (e.g., *equal opportunity difference* [34], n = 18) being the most frequently used, followed by summary measures (e.g. *overall accuracy gap* [79], n = 7), and only one clinical utility metric having been identified (*subgroup net benefit* [88]). Only 15 out of these 47 metrics had been explicitly proposed in healthcare contexts. (Table 1)

A structured "catalogue" of fairness metrics, is summarised in four supplementary tables:

- Supplementary Table 4.1: classification within the fairness taxonomies;
- Supplementary Table 4.2: operationalisation details;
- Supplementary Table 4.3: interpretation and implications;
- Supplementary Table 4.4: justified use and critical appraisal

## Critical Appraisal of the fairness metrics

Based on this catalogue of fairness metrics, we critically appraise them in terms of their applicability, interpretability, quality of definition, validation studies, and alignment with clinical and ethical considerations. Below, we summarise key qualitative findings for each category, providing guidance on which metrics to prioritise when assessing the fairness of clinical prediction models (Table 2).

**1. Performance-Independent Metrics** typically aim to achieve parity in the model's positivity rates.[53] For these metrics, a "perfectly accurate" model (i.e, no classification errors after setting a decision threshold, at the population level) may never achieve complete fairness if prevalence differs across groups defined by the sensitive attribute. [89] (Box 2)

**1.1 Probability-based metrics** (e.g., Mean Score Parity[90]) should be used with caution as they are limited by their reliance on mean values without a defined measure of uncertainty. Metrics such as Unsupervised Ranking Fairness (URF) [56] may have value if some form of "unsupervised discrimination" analysis was deemed necessary, but is inadvisable without further empirical evaluation, as their theoretical foundation and validation in healthcare remains unclear.

**1.2 Threshold-dependent metrics** such as Statistical Parity[60], children metrics [61, 55] , and Disparate Impact[62] can be useful for assessing fairness at the predicted class level, without relying on the quality of existing labels. Conditional Statistical Parity[41] metric is especially relevant where legitimate factors (e.g., comorbidities or biological differences) justify group-specific (i.e., differential) risk estimates. The Disparate Impact metric has been connected with the U.S. disparate impact law[62] (even though the four-fifths rule that is often associated with this metric is neither necessary nor sufficient to abstract the disparate impact law[91]), whereas Conditional Statistical Parity aligns with EU non-discrimination law, which could be desirable from a legal perspective[92]. However, operationalising fairness remains inconsistent when sensitive attributes are non-binary or require conditioning across multiple categorical strata. While survival[58] and censoring-based[59] metrics extend fairness evaluation to more complex scenarios – their adoption is hindered by insufficient empirical evaluation and behavioural understanding. Similarly, metrics such as Differential Fairness[55] lack sufficient empirical evidence and are currently inadvisable for real-world use.

**2. Performance-Dependent Metrics** compare performance across individuals or subgroups of individuals. According to these metrics, a model is perfectly fair when it achieves, for example, consistent (but not necessarily perfect) discrimination, calibration, or clinical utility across subgroups. These metrics assume that the distribution of the predictors and outcomes present in the model development data reflect the target population. By prioritising consistency (i.e., statistical parity across individuals/groups) in model performance across subgroups, the *status quo* in healthcare delivery is preserved and labels are reinforced [53] . Whether aiming for statistical parity should be a priority remains debated: some argue that performance may naturally vary between subgroups and that the focus should instead be on ensuring minimum acceptable performance for all [89]. See Box 3 and 4 for scenarios and examples of performance-dependent fairness metrics applied in the healthcare literature. These can be categorised as follows:

**2.1 Probability-based metrics** assess disparities in model performance across subgroups using base performance metrics that are probability-based, i.e., do not apply thresholds. Such metrics are often considered as preferable for clinical prediction models, and can be further categorised into discrimination, calibration, or overall metrics [1]. (Box 3)

**2.1.1 Discrimination metrics**: following a previous study [1] on these base performance metrics, AUROC Parity [63] is recommended for quantifying discrimination disparities, whereas AUPRC Parity [64] is inadvisable due to its semi-proper scoring nature, lack of focus (mixing discrimination and clinical utility)[1], and being a discriminatory metric that favours higher-prevalence subgroups [93]. Extensions like xAUROC [65], xAUROC disparity [65], sAUROC [66] , Pairwise Ranking Fairness (PRF) [56], Mean Performance-Scaled Disparity (PSD) AUROC [67], Equity-scaled (ES) AUROC [51], Concordance Imparity [70], and others lack a clear motivation, sufficient implementation details, or in-depth clinical validation, making them hard to interpret and inadvisable to use.

**2.1.2 Calibration metrics:** Equal Calibration [72] and Well Calibration [45] are vaguely defined in the literature and should be used with caution while no clear implementation details on how to consider multiple thresholds are further proposed or studied. Nevertheless, such metrics that satisfy the notion of equal calibration across groups are (in principle) desirable [94]. Expected Calibration Error (ECE) Parity [68], Calibration-in-the-Large Parity, Absolute Calibration Error (ACE) Parity [71], and similar measures should also be used with caution. Such metrics can be informative to assess calibration but should be accompanied by a calibration plot for each subgroup and be paired with discrimination metrics. Furthermore, despite being proper as metrics [1], ECE and ACE have been criticised for being dependent on how calibration binning is done and for handling over- and underestimation equally [95]. Behaviour of fairness metrics focusing on differences in ECE or ACE is not well studied.

**2.1.3 Overall metrics:** Brier Score Parity [64] and Log-Loss Parity [71] should be used with caution: despite being proper measures (Box 1) of model performance [1], they offer limited interpretability. They are also limited in the sense that they provide an overall assessment of the model and are influenced by elements of discrimination and calibration [1]. Balance for Positive Class and Balance for Negative Class [72] should also be used with caution because more knowledge of their behaviour in clinical models is lacking. The Earth Mover's Distance (EMD) for equivalent separation [64], which compares the distributions of estimated probabilities per outcome label using the Wasserstein distance, can be conceptually interesting as it seeks to fulfill the "separation non-discrimination criterion of fairness" [44] in its strongest form. Its behaviour is expected to be proper, but it lacks sufficient empirical evaluation.

**2.2 Threshold-dependent metrics** compare classification performance across groups. These metrics are inherently limited by the fact that they rely on a decision threshold that may be chosen with no clinical rationale. As a result, the reported quantity is specific to the chosen threshold and may not generalise to others. Moreover, some of these metrics can be improper on their own [1,93]. These metrics are usually split into partial, summary, and clinical utility metrics. (Box 4)

**2.2.1 Partial metrics:** Equal Opportunity Difference [34] and Predictive Parity [77], which seek TPR and FPR parity, can be relevant if reported together. Similar observations can be made regarding Predictive Parity [77] and NPV Parity [29]. These metrics have the advantage of being easily interpreted. However, their reliance on thresholds that may be somewhat arbitrary limits standalone use. We regard Equalised Odds[34] as inadvisable because combining TPR and FPR makes the metric lose focus, hampering a clear interpretation. Moreover, its behaviour (for example in comparison with an accuracy metric) has not been sufficiently explored. Worst-group Recall [82] and Maximum Difference Recall [80] should be used with caution and can be relevant under Rawlsian fairness principles, in which the benefit of the worst-off subgroup should be maximised. Recall Intergroup Standard Deviation (ISD) [83], Recall Coefficient of Variation [64] , Recall Disparity Ratio [81], and Recall Between-group Generalised Entropy Index (GEI) [84] are overly complex as metrics, their behaviour has not been sufficiently studied, and, as a result, may be hard to interpret. Finally, Recall-HEAL [85] introduces reparative fairness by computing an anticorrelation between model performance and historical disease burden. However, complex tradeoffs and ethical debates may arise, and the use of external data can be deemed arbitrary, which makes the use of this metric (and family of metrics) questionable.

**2.2.2 Summary metrics:** these metrics (e.g., Overall Accuracy Gap [79], Error Rate Ratio [86], Balanced Accuracy Difference [29]) are inadvisable due to improper behaviour (Box 1), lack of distinction of error types, and threshold-dependency. F1 Parity [52] is inadvisable due to improper scoring behavior. MCC Parity [87] is also highly questionable due to its highly complex formulation that hampers interpretability. The Treatment Equality [79] metric has unclear motivation and implications.

**2.2.3 Clinical utility metrics:** only one metric was identified (subgroup net benefit [88]). Although further empirical evaluation is needed, this metric can potentially be particularly relevant [1,96,97] as it reflects the quality of decision-making (at a clinically meaningful threshold), capturing the trade-off between potential harms and benefits. Net benefit, defined as *sensitivity* × *prevalence* – *(1 – specificity)* × *(1 – prevalence)* × *w*, where *w* is the odds at the threshold probability, will be lower with lower prevalence, meaning that subgroups with lower prevalence can only be expected to have a lower prevalence [96,97]. Subgroup net benefit, as defined in [88], accounts for different outcome prevalence across subgroups and is particularly suited when the focus of benefit lies in true negatives. Furthermore, it allows each subgroup to have its own prevalence term for the outcome of interest, which may be relevant in the context of pre-existing prevalence differences or health inequities.

## Discussion

In this scoping review, we reviewed 41 studies and identified 62 distinct fairness metrics, which we categorised based on the taxonomy we proposed. We considered how each metric should be interpreted, in light of applicable ethical and legal frameworks. We then investigated when the use of each metric is justified. Collectively, we offer a qualitative critical appraisal and practical guidance that considers the implications, limitations, and appropriate contexts for each metric, providing broader recommendations for fairness evaluation of clinical prediction models (Box 5).

Our review revealed broader issues that warrant discussion. The definition of "fairness metric" in the literature is not clear or satisfactory (Supplementary Table 2.2). This conceptual ambiguity undermines our ability to systematically assess the fairness of clinical predictive models. To address this gap, we proposed a working definition of "fairness metric" (Box 1), which also serves as the foundation for our methodology. The proliferation of fairness taxonomies further exacerbates this problem. While diversity in methodological approaches can be beneficial, the excessive number of proposed metrics and definitions can result in redundancies and overlaps, with subtle variations in their formulations leading to inconsistent interpretations and implementations.

### Conceptual challenges and risk of "epistemic trespassing"

Fairness metrics largely originate from computer science [36] and are frequently evaluated in case studies outside healthcare, such as recidivism prediction [94] or credit scoring [34]. These metrics often emerge in isolation, or at risk of "epistemic trespassing" [91], without a clear ethical or theoretical foundation [89], raising concerns about their suitability for clinical use [98].

Challenges in operationalising fairness metrics increase as their complexity grows. As soon as fairness assessments move beyond "binary classification tasks", "binary sensitive attributes", or simple conditioning approaches, it is unclear how to compute (and interpret these metrics if someone else computed them).

We found that fairness metrics often lack clear definitions, justified use, and sufficient empirical evaluation, limiting their reliability for clinical predictive AI in real-world scenarios. Many of the metrics suffer from theoretical ambiguities, potential inconsistent behaviour, or inadequate empirical support. The level of scrutiny varies widely — some metrics are rigorously studied and grounded in ethical or legal principles (e.g. equality of opportunity difference [34]), while others are introduced with minimal justification (e.g. xAUROC [65]).

### Methodological patterns in the fairness metrics landscape

The predominance of performance-dependent (46/62) and threshold-dependent (33/62) metrics further underscores a methodological convenience bias — these metrics are easier to compute, but may fail to capture a full and relevant picture of model behaviour. Performance-dependent metrics, for instance, may preserve existing biases in the data rather than correcting them [53], while threshold-dependent metrics may be inherently limited as they assess fairness at specific and arbitrarily chosen decision cut-offs rather than evaluating overall model behaviour [99].

Most of the identified metrics are group fairness metrics (n = 60/62). Individual fairness metrics were notably scarce (n = 2/62), mainly due to the fact that these are hard and sometimes controversial to operationalise as metrics. We suggest interested readers on individual fairness refer to [46], although the authors only found six uses of individual fairness in the healthcare space [59,100–104], and their operationalisation as metrics is unclear. Beyond group and individual fairness, intersectionality presents an important consideration in fairness assessment. Intersectionality refers to *the way overlapping social identities interact to create unique experiences of privilege or oppression* [105]. Although it can be operationalised as group fairness with varying subgroup granularity, "intersectional" or "subgroup" families of fairness metrics have also been proposed [55,106]. Even though existing fairness metrics often assess disparities along single attributes (e.g., sex, age, race), real-world unfairness arises at the intersections of these dimensions: for example., an elderly Black woman may face distinct disadvantages not captured by separate assessments of age, race, or sex [107] .

Another notable gap in the literature was the limited presence of clinical utility metrics. Only one metric, subgroup net benefit [88], was explicitly defined as capturing clinical utility, despite the relevance of such metrics for informing decision-making [1]. Two other metrics — equalising disincentives [74] and treatment equality [79] — appear to be inspired by utility reasoning, but were neither explicitly framed as clinical utility metrics nor originally proposed in a healthcare context. Existing fairness metrics focus largely on parity in statistical attributes but does not expose any downstream impact on patient outcomes, which is arguably the most important for models in a clinical setting. Implementing predictive AI must ultimately translate into improved patient outcomes, therefore future research should prioritise metrics that explicitly link fairness to clinical utility [108]. For instance, subgroup-specific decision curves and net benefit-based fairness metrics could have significant value [109].

## Parity versus minimum acceptable performance

Most fairness metrics are parity-based and primarily capture numeric discrepancies between groups. However, such differences do not necessarily correspond to clinically meaningful disparities. Their interpretation is highly dependent on the performance regime of the model. For instance, a difference in AUROC of 0.90 and 0.80 across subgroups may have very different clinical implications compared to a difference of 0.70 and 0.60. This variability raises important considerations regarding when a disparity should prompt concern, and whether statistical significance alone is a sufficient basis for intervention. Rather than aiming for statistical parity at all costs, fairness should be framed in terms of minimum acceptable performance for all groups [89], guided by clinical relevance and known health disparities.

The downstream impact of fairness violations is inherently context-dependent. Even in scenarios where all groups benefit from the model, the fact that one group benefits less may still be problematic. While this could represent an improvement over standard practice, tolerating such disparities may have longer-term implications. Once deployed, models are subject to data drift and performance degradation. In these cases, groups that initially benefited less may be the first to experience a decline in performance, a phenomenon that has been described as "fairness drift" [110].

## Gaps

We have identified critical gaps in fairness metric development and evaluation that remain unaddressed:

- **Sample size:** what are the sample size requirements to evaluate fairness [111,112];
- **Uncertainty:** most fairness metrics do not provide confidence intervals, which may be crucial as the groups defined by sensitive groups will typically have very different (and often limited) sample sizes [113];
- **Intersectionality:** most fairness assessments treat sensitive attributes independently, overlooking potential interactions between multiple axes of disparity [114]. Considering intersections can be difficult due to small intersectional subgroup sizes even in large datasets [111,115–117].
- **Probability-based fairness metrics:** given the subjective and arbitrary nature of decision thresholds, during model validation fairness metrics should prioritise probability-based fairness metrics (that are not dependent on thresholds);
- **Clinical utility:** there is little work integrating fairness assessments with clinical utility-related metrics, which should be prioritised;
- **Lack of empirical behaviour analysis:** many proposed fairness metrics lack an evaluation of their behaviour in healthcare and across different datasets and settings, making it difficult to assess their reliability and usefulness;
- **Fairness metric tradeoffs:** it is well-known that certain metrics cannot be fulfilled simultaneously [72], yet these conflicts have not been extensively studied in clinical predictive AI.

Future work should prioritise fairness metrics that provide uncertainty estimates, support intersectional analyses, align with clinical outcomes and benefit, and are systematically evaluated across real-world healthcare settings.

## Limitations of this work

Our search strategy did not explicitly include the terms "bias", "parity", or "disparity", which may have led to the omission of relevant studies using different terminologies, despite the long list of metrics we found, which covered metrics identified in previous reviews, as well as new ones. Expanding search terms in future work could improve coverage. We have not conducted an extensive assessment of practical use cases of fairness metrics in the literature, nor have we measured their impact. Additionally, our review primarily focused on a qualitative critical appraisal of fairness metrics, with no quantitative analyses. Future research should translate our qualitative critical appraisal into quantitative evidence, via simulation studies or using real-world data. Finally, future work should focus on developing practical recommendations for selecting and applying fairness metrics in clinical predictive models, proposing methodological solutions to address identified technical gaps, and evaluating the real-world impact of different fairness assessments.

## Conclusion

The current landscape of fairness metrics in clinical predictive AI is fragmented, with a lack of clear definitions, standardisation, and clinical relevance. Several metrics were found to have insufficient empirical evaluation, hampering their relevance in real-world settings. Future research should prioritise fairness metrics (and assessments in general) that align with clinical decision-making, incorporate uncertainty estimation, and account for intersectionality. Fairness evaluations should be conducted contextually and collaboratively with key stakeholders to ensure that they meaningfully contribute to equitable healthcare outcomes.

# Main Exhibits

## Box 1. Glossary of terms used in the review

| Concept | According to | Definition |
|---|---|---|
| Algorithmic Bias | Statistics | Property that a model exhibits when data or analysis biases are encoded directly or indirectly into it |
| Bias | Society | Being biased against a certain individual or group means that such individual or group is consistently disadvantaged. |
| Bias (Systematic Error) | Statistics | Consistent or proportional difference between the predicted value and the observed value; also known as systematic error |
| Bias Preserving Fairness Metrics | Statistics | Fairness metrics that aim to maintain the patterns observed in the data used for model development, such as specialist follow-up referral rates. These metrics assume that the distributions and outcomes present in the data reflect acceptable baselines. As a result, they prioritise consistency in model performance across groups rather than adjusting for underlying biases, effectively preserving the *status quo* in healthcare delivery. [53] |
| Bias Transforming Fairness Metrics | Statistics | Fairness metrics that do not assume that clinical or societal biases should be preserved in model development (e.g, disparities in disease prevalence, access to care, or historical disease patterns). These metrics are typically independent of model performance and instead aim to achieve parity in the model's predicted positivity rates. Their goal is to actively modify the influence of underlying data biases to promote greater fairness in outcomes.[53] |
| Calibration | Statistics | The validity of risk estimates, relating to the agreement between the estimated and observed number of events [99] |
| Clinical Utility | Statistics | Metrics and plots that evaluate the potential benefit of model-guided decisions, by assessing whether such decisions are likely to lead to better outcomes or fewer harms compared to alternative strategies (e.g., standard care, treat-all, treat-none, or another model). |
| Clinical Prediction Model | Statistics | A model that aims to estimate the probability of present or future health outcomes given a set of baseline predictors to facilitate medical decision making and improve people's health outcomes [118] |
| Decision Threshold | Statistics | The probability cut-off at which a model categorises a sample as belonging to a particular outcome class, based on the estimated probability. In clinical settings, this threshold determines, for example, whether an intervention is triggered or a diagnosis is made. |
| Discrimination | Society | Prejudicial treatment of groups of individuals, based on group membership |
| Discrimination | Statistics | How well the predictions from the model differentiate between individuals with (high-risk) and without (low-risk) the outcome |
| Disparate Impact | Society | When a policy or practice has an adverse effect on a protected group (i.e, a category of individuals who are legally safeguarded against discrimination under specific laws and regulations), even though it appears neutral |
| Equality | Society | Everyone should (in principle) get the same treatment, regardless of baseline conditions or potential outcomes. |
| Equity | Society | Equals should (in principle) be treated equally, and unequals unequally; treat |
| | | like cases alike such that everyone attains their full potential. |
| Evaluation (or performance) metric | Statistics | Quantitative measure or plot that assesses the performance or effectiveness of a prediction model's output according to an outcome label (or "ground-truth", or "reference standard") |
| Fairness | Society | Honesty; impartiality, equitableness, justness; fair dealing. [119] |
| Fairness | Statistics | A property of a prediction model whereby individuals or groups defined by protected attributes (e.g., age, race/ethnicity, sex/gender, or socioeconomic status) are not systematically disadvantaged in terms of model outputs or associated decisions. |
| Fairness Metric | Statistics | A measure that quantifies the extent to which a model's output does not discriminate (in the societal sense, based on a given notion of fairness) against individuals or groups defined by a sensitive attribute |
| Fairness Notion | Statistics | Definition of fairness that a model can satisfy or fail to satisfy (e.g counterfactual fairness, according to which a decision is fair if it remains the same regardless of whether an individual belongs to a different group defined by a sensitive attribute [37]) |
| Group Fairness | Statistics | Groups of individuals defined by sensitive attributes should (in principle) receive similar care [44] |
| Hidden Stratification | Statistics | When a model appears to perform well at the population level but exhibits poorer performance in one or more subgroups, potentially leading to disparities in its predictions [32,33] |
| Individual Fairness | Statistics | Similar individuals should (in principle) receive similar care [36]. The concept of "similar" is defined according to a measure or distance that can be adjusted and agreed upon by relevant stakeholders (e.g., patients, healthcare professionals, policy makers). |
| Intersectionality | Society | The interconnected nature of social categorisations such as age, race/ethnicity, sex/gender, or socioeconomic status, regarded as creating overlapping and interdependent systems of discrimination/disadvantage [114] |
| Justice | Society | Conformity (of an action or thing) to moral right, or to reason, truth, or fact[120] |
| Performance dependent metric | Statistics | A fairness metric that compares model performance across individuals or groups of individuals (e.g., *AUROC parity*). See "bias-preserving fairness metrics" entry for implications. |
| Performance independent metric | Statistics | A fairness metric that compares model behaviour (rather than performance, assessing, for example, parity in positivity rates) across individuals or groups of individuals (e.g., statistical parity). See "bias-transforming fairness metrics" entry for implications. |
| Probability based metric | Statistics | A metric in which estimated probabilities are used as input (e.g., *AUROC parity*) [1] |
| Proper measure | Statistics | A performance measure is proper if its expected value is optimized when using the correct probabilities [1] |
| Sensitive (or Protected) Attribute | Statistics | A characteristic, feature, variable, or axis according to which fairness can be evaluated (such as age, race/ethnicity, sex/gender, or socioeconomic status), often derived from legal frameworks |
| Threshold dependent metric | Statistics | A metric in which predicted classes (a result of converting estimated probabilities using a decision threshold) are used as input (e.g, equality of opportunity difference) [1] |

## Table 1. Fairness metrics, description, and number of metrics found in the review

| **Type of Fairness →**<br>↓ Type of Metric [1] | **Description / Rationale** | **Individual Fairness (N)** | **Group Fairness (N)** | **Proposed in Healthcare? (N)** |
|---|---|---|---|---|
| **1. Performance-independent** | Unsupervised: the metric does not compute performance (against an outcome label / ground-truth) | | | |
| **1.1 Probability-based** | Estimated probabilities are used (e.g *mean score parity*) – "regression" | 2 | 6 | 3 |
| **1.2 Threshold-dependent** | Predicted classifications are used (e.g *statistical parity*) – "classification" | 0 | 7 | 0 |
| **1. Total** | | **2** | **13** | **3** |
| **2. Performance-dependent** | Supervised: the metric compares performance across individuals or groups of individuals | | | |
| **2.1 Probability-based** | Estimated probabilities are used | | | |
| 2.1.1 Discrimination | The model should estimate higher probabilities for individuals who experience an event compared to those who do not (e.g *AUROC parity difference*) | 0 | 11 | 4 |
| 2.1.2 Calibration | Estimated probabilities should correspond to observed event proportions (e.g *equal calibration*) | 0 | 5 | 1 |
| 2.1.3 Overall | Estimated probabilities from the model, in [0, 1], should be as close to actual outcomes, in [0, 1] (e.g *Brier score parity difference*) | 0 | 5 | 0 |
| 2.1 Subtotal | | **0** | **21** | **5** |
| **2.2 Threshold-dependent** | Predicted classifications are used | | | |
| 2.2.1 Partial | Individuals should be classified correctly corresponding to their observed outcome, based on partial view of the confusion matrix (e.g *equal opportunity difference*) | 0 | 18 | 6 |
| 2.2.2 Summary | Individuals should be classified correctly corresponding to their observed outcome, based on the whole confusion matrix (e.g *overall accuracy gap*) | 0 | 7 | 3 |
| 2.2.3 Clinical Utility | Classifications should lead to better clinical decisions (e.g *subgroup net benefit*, the only metric found in this category) | 0 | 1 | 1 |
| 2.2 Subtotal | | 0 | 26 | **10** |
| **2. Total** | | **0** | **47** | **15** |
| **Grand Total** | | **2** | **60** | **18** |

Number of individual and group fairness metrics identified, and whether they were proposed for healthcare.

## Figure 1. Decision diagram guiding the selection of fairness metrics in clinical predictive AI, structured according to the taxonomy

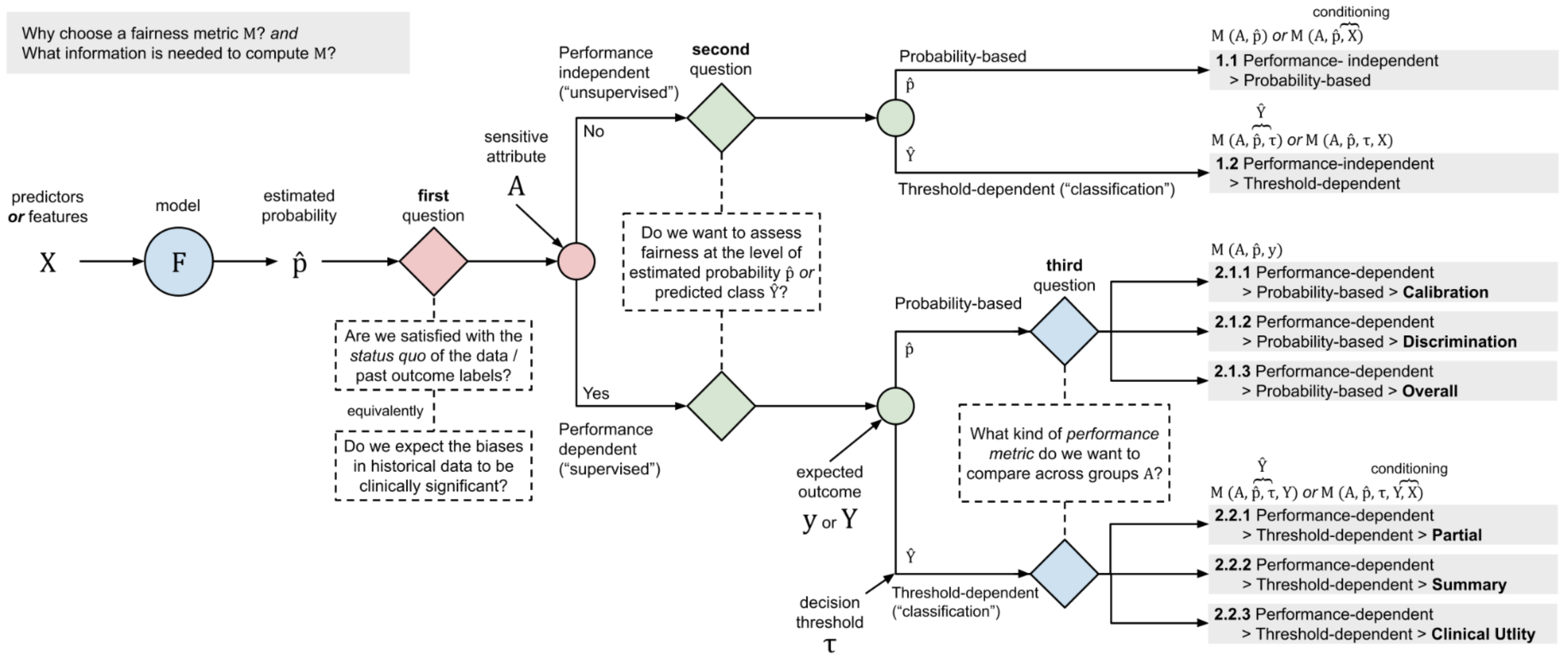

The diagram provides a decision tree to guide the selection of fairness metrics, based on what information is available and what fairness objectives are prioritised. Input features X are processed by a model F, which outputs estimated probabilities $\hat{p}$. Using the sensitive attribute A, outcome labels y, decision thresholds $\tau$, and predicted classes $\hat{Y}$, users navigate three questions, from left to right:

1. Whether the *status quo* or data patterns are acceptable, i.e, whether fairness is assessed independently of labels (“unsupervised”) or in relation to outcomes (“supervised”);
2. Whether the focus is on probability-based or threshold-dependent (“classification”) performance measures;
3. What kind of performance metric is most adequate and we want to compare across individuals or groups?

Each branch leads to a category of fairness metric, classified along two main axes: (i) performance-independent vs. performance-dependent, and (ii) probability-based vs. threshold-dependent. Subtypes of metrics include Calibration, Discrimination, Overall, Partial, Summary, and Clinical Utility. Each category indicates the necessary conditioning variables required to compute the fairness metric.

## Table 2. Guidance and critical appraisal per fairness metric type in clinical prediction models, based on *qualitative* assessment

Three levels of guidance (1 to 3; 1 being the most favourable, and 3 the least): (1) Recommended if; (2) Use with caution; (3) Inadvisable

| Type of Metric | Metrics | Guidance | Comments |
|---|---|---|---|
| **1.1** Performance-independent >**Probability-based** | Mean Score Parity [76], Regression Demographic Parity [57] | Use with caution, provided there are no expected differences in risk across subgroups | Can quickly summarise how a model behaves when comparing different subgroups, in a "bias-transforming" fashion. Limited to the fact that it just reports a mean, and no confidence intervals are usually included |
| | Unsupervised ranking fairness (URF) [56], Survival Group Fairness [58], Censoring-based Group Fairness [59], Survival Intersectional Fairness [58], Survival Individual Fairness [58], Censoring-based Individual Fairness [59] | Inadvisable, unless further validation is conducted and implementation details are clarified | URF can be relevant if some sort of "unsupervised discrimination" is deemed necessary. Not sufficient knowledge or support of such metrics. Some have not been validated in healthcare |
| **1.2** Performance-independent >**Threshold-dependent** | Statistical Parity [60] (and children metrics [61 55]), Disparate Impact [62] | Use with caution, provided positivity rate matters, there are no expected differences in risk across subgroups, and the chosen threshold is relevant | If positivity rate is relevant in the context of the model application and a bias transforming metric is sought. This will be dependent on the selected threshold and therefore context-dependent. Parity achieved with one threshold does not guarantee parity with other thresholds. |
| | Conditional Statistical Parity [41] | Recommended if there is no expected interaction between the legitimate factor and the protected attribute, and there are no "true differences" after accounting for them | Relevant in legal and clinical contexts where consistent positivity rate is important but it may be "legitimately" affected by other factors (e.g comorbidities). Aligned with EU non-discrimination law in light of "contextual equality" [92] |
| | Differential Fairness [55] | Inadvisable, due to lack of validation | Lack of empirical evaluation or significant validation in healthcare, to support the recommendation of such metrics |
| **2.1.1** Performance-dependent > Probability-based > **Discrimination** | AUROC Parity [63] | Recommended if paired with a calibration-related parity metric | Quantifies disparity in discrimination using AUROC. This should be paired with calibration metrics, since AUROC alone is just focused on discrimination. Just like all parity-based metrics, relevance is dependent on the importance of achieving consistent performance across all groups. |
| | AUPRC Parity [64] | Inadvisable, due to metric not being proper and inadequate with varying group sizes | AUPRC is inadvisable by itself [1,93] and it can be an explicitly discriminatory metric through favouring higher-prevalence subgroups [93] |
| | xAUROC [65], xAUROC Disparity [65], sAUROC [66], PFR [56] , PRF Disparity [56] , Equity-scaled AUROC [51], Mean-PSD AUROC [67], Max-PSD AUROC [67], Concordance Imparity [70] | Inadvisable, due to insufficient validation and implementation details. Some could be acceptable after further investigation. | Insufficient knowledge or support of these metrics; seldomly defined or proposed in the healthcare domain. Some of these metrics may be found to be acceptable or relevant after further research and empirical evaluation but they are inadvisable as they stand / with current knowledge. |
| **2.1.2** Performance-dependent > Probability-based > **Calibration** | Equal Calibration [72], Well Calibration [45] | Use with caution, as implementation details lack, and should be paired with a calibration plot per subgroup and a discrimination metric | Unclear how to operationalise across all thresholds, as the metric definition is not sufficiently specific in terms of implementation. Metrics that satisfy the notion of equal calibration across groups would (in principle) be desirable [94] |
| | ECE Parity [68], ACE Parity [71], Calibration-in-the-large Parity [69] | Use with caution, paired with a calibration plot per subgroup and a discrimination metric | Can be informative to assess calibration but only has value when also a calibration plot for each subgroup is also presented, and they should be paired with discrimination metrics. Metrics such as ECE and ACE have been criticised for being dependent on how calibration binning is done and for handling over- and underestimation equally [95]. Behaviour of fairness metrics focusing on differences in ECE or ACE is not well studied. |
| **2.1.3** Performance-dependent > Probability-based **> Overall** | Log-loss Parity [71], Brier Score Parity [64] | Use with caution, and not essential | Both log-loss and Brier score as proper as base metrics, but can be hard to interpret, not as informative due to their "overall" nature that it mixes discrimination and calibration |
| | Earth Mover's Distance (EMD) for equivalent separation [64], Balance for Positive Class [72], Balance for Negative Class [72] | Use with caution, but hard to interpret and lacking more validation | Relevant metrics, in principle, but lacking empirical evaluation of their behaviour, although some have been used in healthcare. Potentially hard to interpret, especially EMD |
| **2.2.1** Performance-independent >Threshold-dependent **> Partial** | Equal Opportunity Difference [34], Recall Equality Difference [80], Predictive Equality [73] | Use with caution, provided they are reported together, and if TPR, TNR are relevant | The choice of a decision threshold is often arbitrary. If the threshold is being appraised, reporting these metrics together can be descriptively informative. Nevertheless, the base metrics are improper on their own [1,93]. |
| | Predictive Parity [77], NPV Difference [29] | Use with caution, provided they are reported together, and if PPV, NPV are relevant | *Id.* cell above. PPV and NPV can be more practical measures as the condition is on the classification. The base metrics are also improper on their own [1,93]. |
| | Equalised Odds [34], Equalising Disincentives [74] , Average Odds Difference [75], Average Disparity in Equalised Odds [76], Conditional EO Gap [78], | Inadvisable, despite being widely used in the literature | Unclear how (TPR and FPR), or (PPV and NPV) should be combined, with aggregation of both quantities not always well defined; added value compared to separate reporting is |
| | Conditional Use Accuracy Equality [79] | | not clear; naming conventions can be misleading (e.g “odds”). These metrics are often incompatible with each other; e.g., if prevalence differs across subgroups, equalised odds and predictive parity cannot be achieved simultaneously [121] |
| | Worst-group Recall [82], Maximum Difference Recall [80] | Use with caution, descriptively, if Rawlsian fairness notions are sought and outliers are not expected | Linked to Rawls' *Theory of Justice* where the minimum benefit should be maximised across subgroups; can be a strict fairness condition with consequences on levelling down [89]. This is sensitive to outliers, e.g. if a single group with poor performance (e.g due to sample size) exists. |
| | Recall ISD [83], Recall Coefficient of Variation [64], Recall Disparity Ratio [81], Recall Between-group Generalised Entropy Index (GEI)[84] | Inadvisable due to over-complexity | Justified use is unclear and expected behaviour has not been sufficiently explored; as a result, hard to interpret |
| | Recall-HEAL [85] | Use with caution, but further validation and debate are necessary | Relates to reparations theory, which society may want to promote. Complex tradeoffs and ethical debates may arise. The use of external data can be deemed arbitrary |
| **2.2.2** Performance-independent >Threshold-dependent > **Summary** | Overall Accuracy Gap [79], Error Rate Ratio [86], Balanced Accuracy Difference [29], F1 Parity [52], MCC Parity [87], Error Distribution Disparity Index [76] | Inadvisable due to improper behaviour and threshold-dependency | Flawed models may yield higher values than correct ones; all errors are treated in the same way; some are hard to interpret; are limited to one clinical decision threshold. [1,93] |
| | Treatment Equality [79] | Inadvisable | Seems to be utility-inspired but not explicitly defined as such. Motivation and implications are unclear |
| **2.2.3** Performance-independent >Threshold-dependent > **Clinical Utility** | Subgroup net benefit [88] | Recommended if the focus of benefit is true negatives | The most relevant category of metric, as it combines discrimination and calibration [97]. For this specific formulation, the focus of benefit is on true negatives. The formulation assumes different prevalence per subgroup[88]. Further empirical evaluation should be conducted. |

***Box 2: scenarios and examples of performance-independent fairness metrics applied in the healthcare literature***

**Scenario: performance-independent metrics**

To illustrate this, consider Obermeyer and colleagues' work, in which a healthcare model exhibited racial bias due to the use of healthcare costs as a proxy label to estimate future health needs[22]. Because access to healthcare services had historically been lower among Black patients, this label systematically underestimated the true health burden in this population. Although the model achieved high overall performance, the underlying bias went undetected until after deployment. Had a performance-independent metric (i.e., "bias-transforming") been used during development, this disparity may have been identified earlier, highlighting the limitations of relying solely on conventional performance metrics.

**Example: Threshold-dependent metrics**

Ravindranath et al. describe various approaches for developing an XGBoost-based clinical prediction model to estimate whether patients with glaucoma will require incisional glaucoma surgery within 12 months. The data used to develop the model was electronic health record (EHR) data from nearly 40,000 patients across seven US health systems. The study used demographic parity (called "independence" in the paper) as one of several fairness metrics to measure whether the model's prediction rates were equal across different demographic groups regardless of actual outcomes. Among their findings, models that excluded sensitive attributes were less fair than models that included sensitive attributes (e.g., with respect to sex, demographic parity of 0.134 versus 0.038, respectively; see Supplementary Table S5).[122]

***Box 3: scenarios and examples of performance-dependent probability-based fairness metrics applied in the healthcare literature***

**Scenario: performance-dependent metrics**
Consider a clinical prediction model used to guide prostate biopsy decisions. In this context, men with elevated prostate-specific antigen (PSA) levels, typically 3 ng/mL or higher, may be referred for biopsy, but only a small proportion are ultimately diagnosed with high-grade cancer. Biopsy results, which serve as the "ground-truth", are considered reliable in identifying the presence and aggressiveness of cancer. Prediction models have been developed to improve risk stratification, recommending biopsy only when there is a high predicted probability of clinically significant disease.[96,97] This can be viewed as a justified application of "bias-preserving" metrics, given that the outcome is based on a well-established and clinically meaningful reference standard. However, while this may be appropriate for this specific use case, it is important to recognise that such assessments may overlook disparities in access to follow-up testing or subpopulation-level differences in PSA kinetics. Even in these circumstances, a careful use of stratification and performance-independent metrics such as conditional statistical parity can be a useful diagnostic aid. Differences in positivity rate between subpopulations should be explained by difference in known risk factors between the groups; where this is not the case it can indicate the importance of a previous unconsidered risk factor, or some form of sampling selection bias in the data acquisition process.

**Example: Discrimination metrics**
Byrd and colleagues evaluated the performance of Epic's proprietary Deterioration Index (DTI), a prognostic clinical prediction model that estimates the risk of clinical deterioration in hospitalized patients. The study analyzed over 5 million DTI predictions for 13,737 patients across 8 Midwestern US hospitals, defining deterioration as mechanical ventilation, intensive care unit (ICU) transfer, or death. The researchers used AUROC parity as a key bias metric to assess whether the model performed equally well across different demographic subgroups. In contrast to the difference-based AUROC parity metric defined in our review, AUROC parity in this study was calculated as the ratio of the AUROC for a protected group to that of a reference group, with values closer to 1.0 indicating similar discriminative performance. The findings revealed variable performance across demographic groups, with AUROC parity higher than 1.00 for most groups except those who chose not to disclose their ethnicity (0.93).[123]

**Example: Calibration metrics**
Pfohl and colleagues describe the development of a clinical prediction model for atherosclerotic cardiovascular disease risk using EHR data from over 250,000 patients. The researchers employed adversarial learning techniques to create a model that ensures similar error rates (i.e, proportion of classification errors after setting a decision threshold) across different demographic groups defined by race, sex, and age. Among the metrics used to assess fairness, the authors included ACE parity, which quantifies differences in calibration between subgroups. ACE parity assesses whether models consistently over- or underestimated risk across groups[71]. Calibration is particularly important in this setting, as clinical guidelines recommend initiating interventions based on fixed risk thresholds of 7.5% and 20%, which define intermediate- and high-risk categories, respectively[124].

**Example: Overall metrics**
Beyond calibration-focused metrics such as ACE parity, the study by Pfohl and colleagues (mentioned in 2.1.2 Calibration metrics) also considered overall fairness measures. Specifically, to evaluate how well their models aligned the distribution of risk predictions across groups, they used EMD (for equivalent separation), which quantifies differences between the distributions of predicted ASCVD risk across subgroups of patients, conditioned on the true outcome. Their results demonstrated that the adversarial training approach resulted in a substantial reduction in the mean pairwise EMD between each predictive distribution in both outcome strata for both gender and age, with a negligible effect for race. For example, for the positive outcome, with respect to gender, a standard modelling approach achieved an EMD of 0.0167, whereas the "improved model" achieved an EMD of 0.00593 (see Table 2 of the article), which translates to a fairness improvement according to this metric[64].

***Box 4: scenarios and examples of performance-dependent, threshold-dependent fairness metrics applied in the healthcare literature***

**Example: Partial metrics**

Yang and colleagues developed several clinical prediction models for prognosis of three key outcomes in ICU patients: in-hospital mortality, 30-day readmission, and one-year mortality. The researchers linked the publicly-available MIMIC-IV EHR database[125] with community-level social determinants of health (SDoH) data to assess whether the inclusion of SDoH features would enhance predictive performance. They trained multiple model types, including XGBoost classifiers, across various feature sets and patient subgroups. To evaluate algorithmic bias across demographic groups, the study employed predictive parity (FPR parity). For demonstration purposes, a decision threshold of 0.5 was used to convert estimated probabilities into binary classifications. FPR parity indicates whether the model disproportionately flags certain groups as "high-risk." The analysis revealed that patients who were older, Black, female, or from communities with lower incomes, higher public transit usage, or lower educational attainment experienced higher false positive rates and thus lower predictive parity.[126]

**Example: Summary metrics**

Davis and colleagues describe the development of random forest models predicting three post-surgical outcomes (30-day mortality, unplanned readmission, and pneumonia) among veterans receiving surgical care at US Department of Veterans Affairs (VA) facilities. These models were built using features based on the American College of Surgeons NSQIP universal risk calculator on data from 2013, then evaluated for performance drift over a 10-year period (2014-2023). The researchers focused specifically on "fairness drift" (i.e., how algorithmic biases might emerge over time) by measuring, among other fairness metrics, the "accuracy gap" between demographic groups. Classification thresholds for each model were based on the predicted probability that maximized the mean of the specificity and sensitivity in the entire population. This accuracy gap metric quantified the difference in classification accuracy between disadvantaged versus advantaged groups (Black vs White patients, and female vs male patients). The study found that, for instance, the pneumonia model showed increasing accuracy gaps between Black and White patients over time, with Black patients experiencing larger performance declines (see Figure 3 of the article)[110].

**Example: Clinical utility**

Benitez-Aurioles and colleagues extend the concept of net benefit to evaluate how models distribute clinical utility across different subgroups. To showcase their approach, they develop three logistic regression models predicting 5-year incidence of type-2 diabetes to assist clinicians in referring patients to lifestyle intervention programs. The models were built on a UK Biobank dataset of 477,558 patients using relevant demographic and clinical predictors: (1) a model excluding ethnicity as a predictor (LogNoSA), (2) a model including ethnicity (LogSingleSA), and (3) an ensemble of models trained separately for each ethnic group with propensity score weighting (LogMultiSA). The researchers set a clinical threshold of 15% risk (based on the Leicester Diabetes Risk Score used in clinical practice) and a treatment weight of 0.58 (corresponding to the reported relative risk reduction of a diabetes prevention program). They assessed fairness by comparing "subgroup net benefit" (sNB) across ethnic groups, where sNB quantifies the clinical utility of a model in terms of true negatives per 10,000 patients. Their analysis revealed that models including ethnicity as a predictor (LogSingleSA) improved the net benefit for Asian (9,435) and Black (9,597) populations compared to models excluding ethnicity (see Figure 2 of the article) [88].

***Box 5. Key considerations and recommendations for fairness evaluation in clinical prediction models***

- Evaluating model performance at the population level will mask unfairness. Defining subgroups for evaluation by clinically relevant **sensitive attributes** is therefore a fundamental first step to any fairness assessment.
- Although fairness metrics can be valuable for summarising potential biases in model outputs, fairness evaluation is not limited to our definition of fairness metric:
  - **Subgroup performance evaluation**, while not necessarily labelled as a fairness metric, remains essential for identifying groups where the model is underperforming;
  - Metrics related to **discrimination**, **calibration**, and **clinical utility** should be prioritised. Fairness evaluations must be grounded in the goal of improving patient outcomes, which can be measured using clinical utility metrics, arguably the most important for models in a clinical setting;
  - Visual tools such as **calibration plots** and **decision curve analysis**, when stratified by sensitive attributes, are fundamental to highlight performance differences and guide fairness evaluations;
  - In cases where the validity of the outcome labels may be in question, **performance-independent fairness metrics** can help challenge and interrogate the *status quo* embedded in the past data used for model development.
- Performance and model behaviour will vary across subgroups. Rather than aiming for statistical parity at all costs, fairness should be framed in terms of **minimum acceptable performance** for all groups, guided by clinical relevance and known health disparities.
- When clinically defined thresholds exist (e.g. from clinical guidelines), threshold-dependent metrics such as sensitivity **may be** informative descriptively when correctly paired (e.g, sensitivity and specificity together, or PPV and NPV together). However, their interpretation should be constrained to that specific threshold; extrapolating beyond it may result in misleading conclusions. Clinical utility metrics are the only threshold-dependent metrics that are advised on their own.
- **Transparent reporting** is essential, especially given the complexity and plurality of fairness definitions. In addition to adhering to TRIPOD+AI , we recommend explicitly defining and justifying all fairness-related design decisions, metrics, and visualisations.
- Fairness should be defined through **interdisciplinary dialogue**. Proposed metrics should involve diverse stakeholders (e.g, patients, clinicians, caregivers, policymakers) whose backgrounds reflect the populations affected. It is unacceptable for a small, unrepresentative group to unilaterally define fairness for all.

# Supplementary Materials

**Supplementary Material 1:** Data extraction forms
https://osf.io/w9c87

**Supplementary Material 2:**Scoping review extra materials
https://osf.io/qv853

**Supplementary Material 3:**Formal definitions of identified fairness metrics
https://osf.io/4uxty

**Supplementary Material 4:** Catalogue of fairness metrics
https://osf.io/hfsqr